\documentclass{article}
\usepackage{spconf,amsmath,graphicx}
\usepackage{amsfonts}

\usepackage{algorithm}
\usepackage{algorithmic}
\usepackage{epsfig}
\usepackage{multirow}
\usepackage{multicol}
\usepackage{booktabs}
\usepackage{siunitx}
\newcolumntype{P}[1]{>{\centering\arraybackslash}p{#1}}
\usepackage{placeins}
\usepackage{amsmath}
\usepackage{amssymb}

\DeclareMathOperator*{\argmin}{argmin}
\usepackage{pifont}
\usepackage{xcolor}
\usepackage{enumitem}

\title{Single Cell Training on Architecture Search for Image Denoising}
\name{Bokyeung Lee, Kyungdeuk Ko, Jonghwan Hong and Hanseok Ko}
\address{Korea University, Seoul, Korea}

\begin{document}

\maketitle

\begin{abstract}
Neural Architecture Search (NAS) for automatically finding the optimal network architecture has shown some success with competitive performances in various computer vision tasks. However, NAS in general requires a tremendous amount of computations. Thus reducing computational cost has emerged as an important issue. Most of the attempts so far has been based on manual approaches, and often the architectures developed from such efforts dwell in the balance of the network optimality and the search cost. Additionally, recent NAS methods for image restoration generally do not consider dynamic operations that may transform dimensions of feature maps because of the dimensionality mismatch in tensor calculations. This can greatly limit NAS in its search for optimal network structure. To address these issues, we re-frame the optimal search problem by focusing at component block level. From previous work, it's been shown that an effective denoising block can be connected in series to further improve the network performance. By focusing at block level, the search space of reinforcement learning becomes significantly smaller and evaluation process can be conducted more rapidly. In addition, we integrate an innovative dimension matching modules for dealing with spatial and channel-wise mismatch that may occur in the optimal design search. This allows much flexibility in optimal network search within the cell block. With these modules, then we employ reinforcement learning in search of an optimal image denoising network at a module level. Computational efficiency of our proposed Denoising Prior Neural Architecture Search (DPNAS) was demonstrated by having it complete an optimal architecture search for an image restoration task by just one day with a single GPU.
\end{abstract}

\section{Introduction}
Image restoration, a low-level vision task, is aimed to estimate clean images from degraded images. Image restoration problem is usually expressed as $y=\Phi x+n$, where $y$ is a degraded image, $x$ is the original image, $\Phi$ represents the degradation process, and $n$ stands for additive noise. It is a typical ill-posed problem due to the irreversible nature of the image degradation process. Some of the image restoration tasks include image denoising~\cite{BM3D, KSVD, DnCNN, FFDNET, memnet, tvg} and super-resolution~\cite{srcnn, VDSR, EDSR,sparseprior,fbrnn, unsupervised}.

Traditional image restoration methods generally focus on modeling natural image priors and solve as 

\begin{equation} \label{eq1}
x=\argmin_{x}||y-\Phi x||^{2}_{2} + \lambda J(x),
\end{equation}
where \text{$J(x)$} is the regularizer which denotes prior related to \text{$x$}, and \text{$\lambda$} is regularization parameter of \text{$J(x)$}. The many priors have been considered such as sparsity \cite{KSVD, FISTA}, non-local similarity \cite{nonlocal, BM3D}, and gaussian mixture model \cite{GMM}. These prior-based image restoration algorithms can be solved typically by optimization techniques. 

\begin{figure}[t] 
  \centering
  \includegraphics[width=0.5\textwidth]{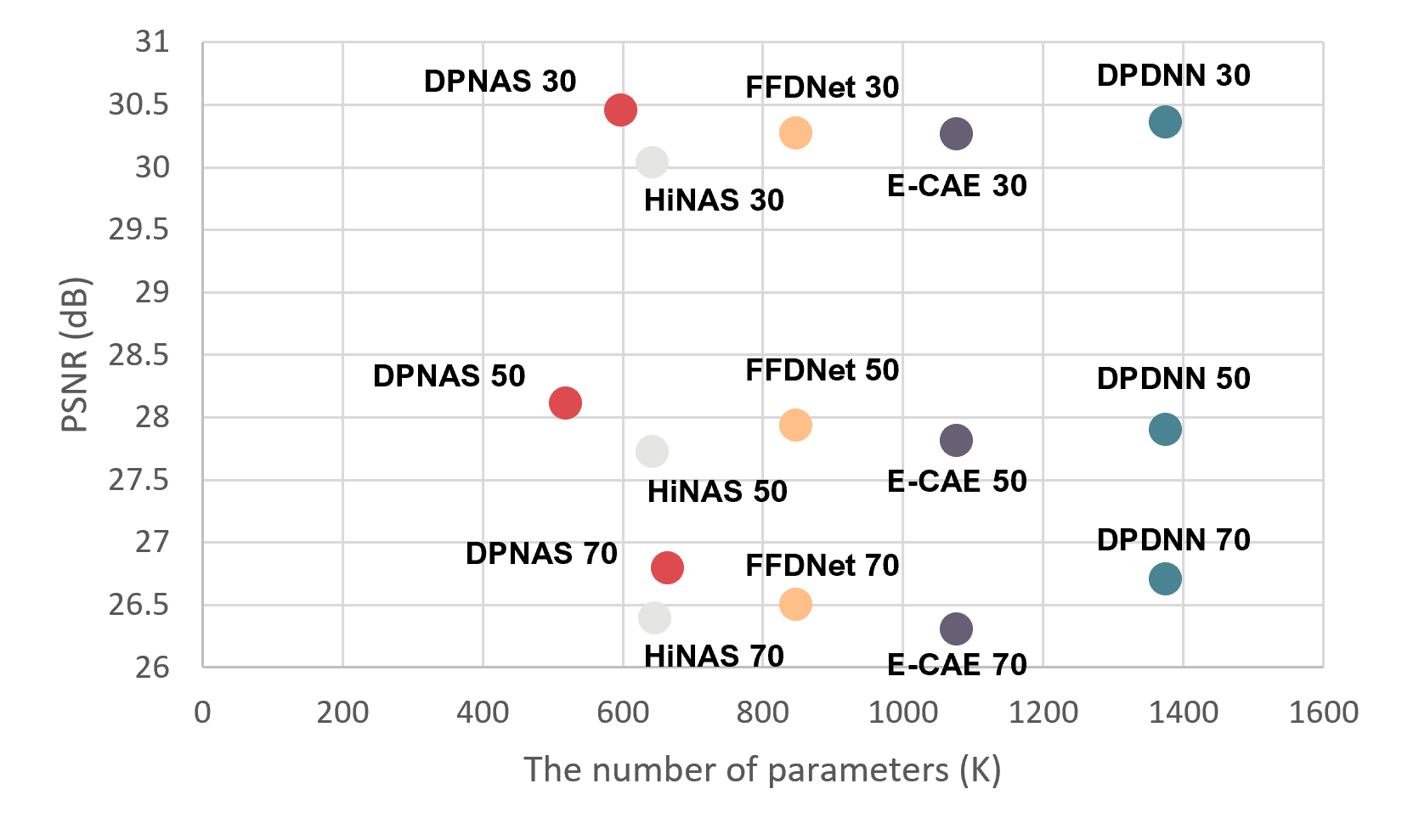}
  \caption{The number of parameters and BSD68 denoising performance comparison with our DPNAS models and other denoising algorithms. K denotes Kilo (\text{$\times 10^3$}). }
  \label{params}
\end{figure}

While there has been an extensive body of works for solving Eq.~\eqref{eq1} considering a variety of degradation processes, deep learning based methodologies have become a dominant form for certain types of degradation processes over the last decade. These relatively recent algorithms with a variety of neural architectures have shown some impressive abilities of estimating original images with remarkable details. There have been some efforts focused on combining both optimization-based structure and deep networks for prior~\cite{ADMMnet, ISTA, denoisingprior, usrnet, featuresparsecoding}. Unlike conventional iterative methods, these learning-based optimization methods consisting of a fixed number of modules are faster and outperform deep learning only based algorithms with fewer trainable parameters. As shown in Eq.~\eqref{eq1}, the role of regularizing the process involved in these neural architectures becomes crucial in acquiring clean images.

However, designing effective network architectures suitable for a given task is not trivial, and often requires extensive effort in identifying a desirable form that can be optimized in meeting task objectives~\cite{selfguided, ridnet, residual, residualnonlocal}. Motivated by this, a growing interest is to automate the model designing process via Neural Architecture Search (NAS)~\cite{NAS}. NAS approaches identify optimum neural structures as a whole network ~\cite{NAS, proxylessnas, mnasnet} or a cell network structure \cite{sharingnas, darts, practical, nat, HINAS}.

However, these NAS methods suffer from three limitations. First, some of the most efficient NASs focused on designing optimal component cell structures while the outer architecture was relied on manual design or implementing existing deep network architectures, i.e. Resnet~\cite{residuallearning}, RDN~\cite{RDN}. Thus the optimization is directed on the cell structure, not on the remaining architecture. Second, the search cost for finding an optimal model architecture is expensive. While the search objective is to find the optimal cell structure, the process requires training the entire model to evaluate cell structure on the validation set. Although various efficient NAS methods have reduced search space, training and evaluating a candidate cell is still time consuming and generally requires enormous computing resources i.e. Block-QNN~\cite{practical} spends 3 days to search a block architecture with 32 GPUs. Lastly, Convolutional Neural Networks (CNN) operations integrated for improving task performance involves downsampling or upsampling, resulting changes in feature dimensions. To avoid dimensional mismatch, Block-QNN avoided operations involving changes in feature map dimensions, and CGP-CNN~\cite{DMM1} pad the outside of input feature maps with zero values before the operation. It doesn't generate various network architecture as well as not suitable for image denoising task.

We address these issues in the following way. First, we propose a novel NAS based on reliable optimization structure for image denoising named Denoising Prior Neural Architecture Search (DPNAS). Secondly, it's been shown by previous work that a well-formed denoising block can improve the performance of the overall model~\cite{DAMP, LDAMP, denoisingprior}. By focusing at cell structure level only, we make it possible not to train the entire model for cell structure evaluation as well as the search space for an optimal denoising architecture becomes significantly smaller. Lastly, To address the issue of dimensionality mismatch from our search space, we developed a novel set of algorithms within DPNAS by employing Dimension Matching Module (DMM) ensure dimensionality matching in CNN operations within the cell structure. Moreover, we propose the search space for image denoising to generate effective denoising network architecture. This allows a significant flexibility in cell structure search space without the danger of integrating CNN operations that may result mismatching tensor dimensions. We summarize the contributions of this work as follows:

\begin{itemize}
\item We developed a novel NAS for image denoising based on search for a component cell structure which is efficacious in optimizing the overall architecture.
\item We developed a search space containing various operations that are dimensionally changeable for noise removal.
\item We developed a dimension matching module which allows flexible combination of CNN operations within the cell structure by a novel set of algorithms enforcing feature dimension matching.
\item The proposed algorithm has shown a remarkable efficiency in that it takes only 1 day for searching one block architecture with one GPU while delivering state-of-the-art performance in image denoising task.
\end{itemize}

\section{Related Work}
The image restoration and reconstruction researches, which have same purpose solving Eq.~\eqref{eq1} with combining deep network and optimization approaches, have been proposed. \cite{ISTANET, featuresparsecoding} combined Iterative Shrinkage and Thresholding Algorithm (ISTA)~\cite{ISTA} and CNN, and achieve efficient image estimation performance using sparse prior knowledge. \cite{ADMMnet} and \cite{denoisingprior} used Alternating Direction Method of Mulpliers (ADMM), which is a popular optimization method, and have outperformed deep learning-based state-of-the-art methods. In particular, \cite{denoisingprior} develops an efficient image restoration method using denoising prior with CNN through several steps. By introducing an auxiliary variable \text{$v$}, Eq.~\eqref{eq1} can be rewritten as 
\begin{equation} \label{eq2}
x,v=\argmin_{x,v}||y-\Phi x||^2_2 + \lambda J(v), s.t. x=v.
\end{equation}
The constrained optimization problem Eq.~\eqref{eq2} can be converted into alternatively solving sub-problem by adopting ADMM, as
\begin{equation} \label{eq4}
\begin{aligned} 
x^{(k+1)} &= \argmin_{x}||y-\Phi x||^2_2+\eta||x-v^{(k)}||^2_2, \\
v^{(k+1)} &= \argmin_{v}\eta||x^{(k+1)}-v||^2_2 + \lambda J(v),
\end{aligned}
\end{equation}
where \text{$k$} is iteration number. Although \text{$x$}-subproblem can be solved in closed-form, it is generally impossible to compute inverse matrix in image restoration task. The authors of \cite{denoisingprior} acquire proximity solution of Eq.~\eqref{eq4} by taking single step of gradient descent, as
\begin{equation} \label{eq5}
\begin{aligned} 
x^{(k+1)} &= x^{(k)} - \delta(\Phi^\mathrm{T}(\Phi x^{(k)}-y)+\eta(x^{(k)}-v^{(k)})), \\
&= \Hat{\Phi} x^{(k)} + \delta \Phi^\mathrm{T} y + \delta\eta v^{(k)}, \\
\end{aligned} 
\end{equation}
where \text{$\Hat{\Phi} = ((1-\delta\eta)I-\delta \Phi^\mathrm{T}\Phi)$} and \text{$\delta$} is the step size. The \text{$v$}-subproblem is a proximity operator of \text{$J(v)$} computed at point \text{$x^{(k+1)}$}, and is considered immediate denoised result as
\begin{equation} \label{eq6}
v^{(k+1)}=f(x^{(k+1)}),
\end{equation}
where \text{$f(\cdot)$} is denoiser. In \cite{denoisingprior}, deep convolution neural network, which is similar to the U-net~\cite{unet}, is used as denoiser \text{$f(\cdot)$}, and \text{$\delta$} and \text{$\eta$} are set to trainable parameters. After \text{$x$} is initialized as \text{$x^{(0)}=\Phi^\mathrm{T} y$}, the restored image can be estimated by iteratively updating two steps Eq.~\eqref{eq6} and Eq.~\eqref{eq5}. The better the denoiser is used, the better the performance of the entire model. Hence, if we find the structure of the optimal denoising architecture, we can maximize the performance and efficiency of the model. 

\begin{table}[t]
\begin{center}
{\scriptsize
\begin{tabular}{|c|c|c|c|c|c|}
\hline
Name & Index & Type & Kernel Size & Pred1 & Pred2 \\
\hline\hline
Convolution & $\boldsymbol{l}$ & 1 & 1, 3 & $\boldsymbol{p}$ & 0 \\
\hline
Downsampling & $\boldsymbol{l}$ & 2 & 2 & $\boldsymbol{p}$ & 0 \\
\hline
Upsampling & $\boldsymbol{l}$ & 3 & 2 & $\boldsymbol{p}$ & 0 \\
\hline
Identity & $\boldsymbol{l}$ & 4 & 0 & $\boldsymbol{p}$ & 0 \\
\hline
Elemental Add & $\boldsymbol{l}$ & 5 & 0 & $\boldsymbol{p}$ & $\boldsymbol{p}$ \\
\hline
Concat & $\boldsymbol{l}$ & 6 & 0 & $\boldsymbol{p}$ & $\boldsymbol{p}$ \\
\hline
Terminal1 & $\boldsymbol{l}$ & 7 & 0 & 0 & 0 \\
\hline
Terminal2 & $\boldsymbol{l}$ & 8 & 0 & 0 & 0 \\

\hline
\end{tabular}
}
\end{center}
\caption{Network Structure Code space for image denoising. The space composed of 8 types that are frequently used for image restoration. $\boldsymbol{l}$ is the layer index 1 to max layer index. $\boldsymbol{p}$ is the predecessor layer indexes 1 to current layer index -1.}
 \label{table1}
\end{table}

\subsection{Deep networks for image restoration}
Deep learning has enjoyed immense success as a key tool for improving performance in a computer vision task. Especially in the image restoration field, various deep network structures for denoising and super-resolution have been recently proposed to improve performance and efficiency. DnCNN~\cite{DnCNN}, IrCNN~\cite{IrCNN}, SRFBN~\cite{srfbn} and GMFN~\cite{gmfn} estimated residual image by adding observation image. NLRN~\cite{nonlocal} and RNAN~\cite{residualnonlocal} employed non-local operation \cite{nonlocal, nonlocalneuralnet} to take wide positions into consideration at time. EDSR~\cite{EDSR} improves the performance of super-resolution by expanding the channel of the feature than the existing super-resolution model. The RED~\cite{red} and SGN~\cite{selfguided} employ low spatial resolution features to extract large scale information and to eliminate redundant elements by using downsamplings such as convolution and pixel-unshuffle.

\begin{figure}[t]
  \centering
  \includegraphics[width=0.5\textwidth]{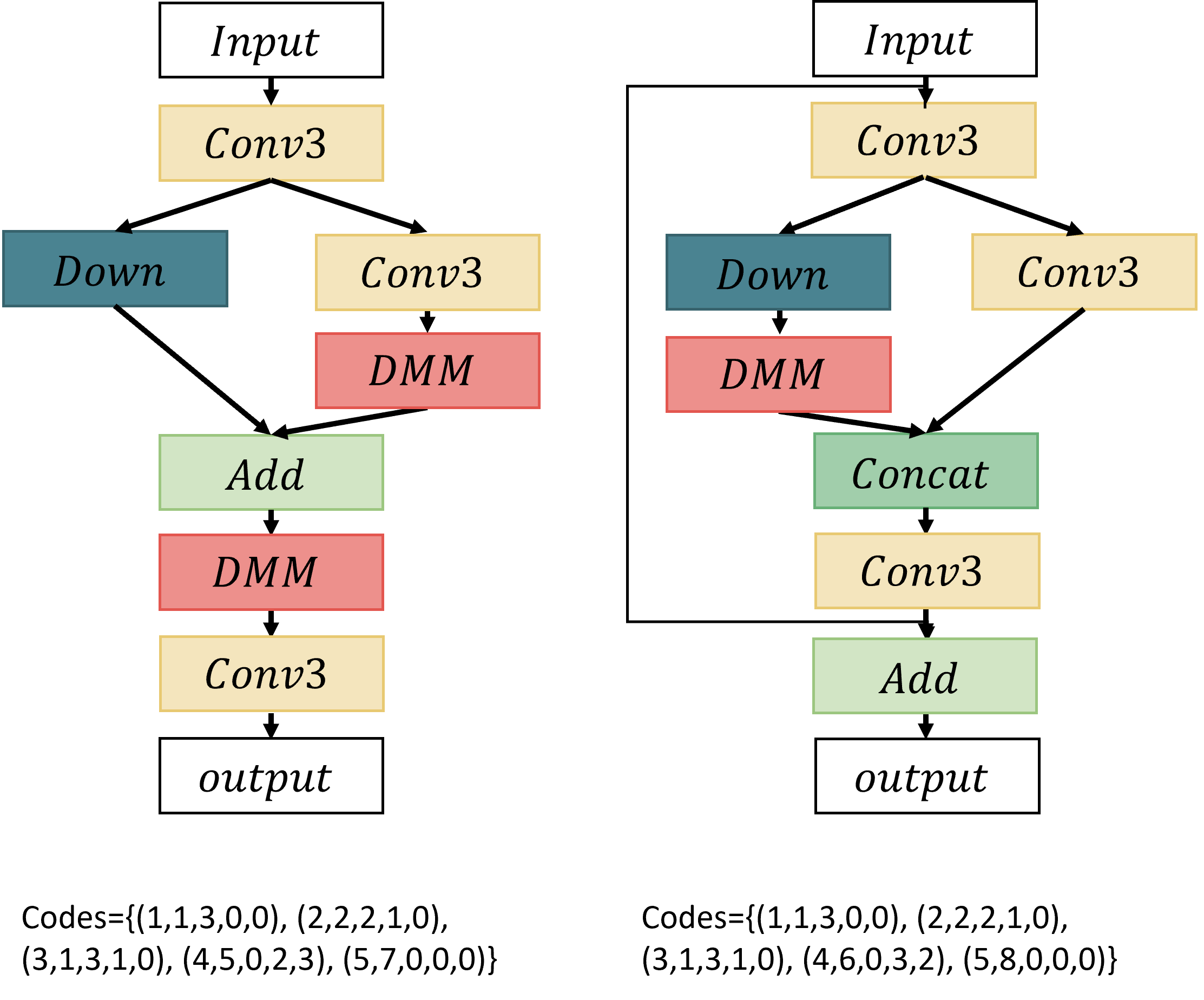}
  \caption{The two examples of the block architecture according to NSC, respectively.} \label{fig2}
\end{figure}

\subsection{Network architecture search (NAS)}

The purpose of NAS is to discover deep neural architecture with high-performance according to the desired application and given datasets, automatically. The representative NAS algorithms, such as evolutionary algorithm, reinforcement learning technique, and DARTS~\cite{darts}-based method, recently have been proposed and achieved competitive performance compared to state-of-the-art methods. Evolutionary algorithm optimizes neural architectures and parameters by iteratively mutating a population of candidate architectures~\cite{EAhi}. Reinforcement learning-based NAS algorithms design the network as sequences from a predefined search space~\cite{NAS}. And, HiNAS~\cite{HINAS}, which is a DARTS-based image restoration method, searches the cell structure on predefined super-cell using gradient descent, which was considered by experienced experts. However, these NAS methods generally require a large number of computations to find the whole network architecture. In contrast, efficient NASs used structures involving human experience in exchange to reduce the search space, which can lead to an inability to cope with various data sets and applications.

\section{Denoising Prior-based Neural Architecture Search}

\subsection{Network search space for image denoising}

Following Block-QNN, we employed Network Structure Code (NSC) as only layer representation. Our search space is designed for image denoising, unlike conventional methods as shown in Table~\ref{table1}. The denoising block is depicted by a set of NSC vectors. The representations of our NSC vectors are similar to conventional one, \textbf{Index}, \textbf{Type} and \textbf{Kernel Size} denote the layer index, operation type and kernel size, respectively. \textbf{Pred1} and \textbf{Pred2} are the index of predecessor. The types of the operation that require one input use only \textbf{Pred1}, and the types of the operation that require two inputs use both \textbf{Pred1} and \textbf{Pred2}. We used PReLU~\cite{prelu} as the activation function following \textbf{Convolution} operation with output channel 64 and stride 1. It led to reducing search space than that with two components separate search.


As downsampling and upsampling layers deliver significant improvement on the image denoising performance, state-of-the-art methods generally employ them. However, these operations are not used in most NAS approaches as NAS has had difficulties in using operations which causes the shape of the input and output to differ. Different from them we employed downsampling layer and upsampling layers to generate flexible network architecture for image denoising. \textbf{Downsampling} operation contains pixel-unshuffle and \text{$1\times1$} convolution, which reduces the spatial size in half and compresses the size of the channel to quarter to maintain the size of the channel. \textbf{Upsampling} operation consists of \text{$1\times1$} convolution and pixel-shuffle, which expands the spatial size by 2 times maintaining channel size. \textbf{Identity} outputs the feature of \textbf{Pred1} without any operation, but is needed to design efficient network architecture.

In aspect to network architecture design, generally, element-wise addition and channel-wise concatenation are considered as operation satisfied associative property, (i.e., add(\textbf{Pred1}, \textbf{Pred2}) == add(\textbf{Pred2}, \textbf{Pred1})). In our case, however, dimension of output of each layer can be different due to various operation. Unlike conventional NSC, we assigned meaning to each \textbf{Pred1} and \textbf{Pred2} in operations that require two inputs. \textbf{Elemental Add} add the features of \textbf{Pred1} to the one of \textbf{Pred2}, which is reshaped by Dimension Matching Module (DMM), it will be described in \textbf{Dimension matching module} section. \textbf{Concat} operations conduct channel wise concatenation for tensor of \textbf{Pred1} and the feature with spatial size of \textbf{Pred1} by adjusting tensor of \textbf{Pred2}. In our search space, \textbf{Elemental Add} and \textbf{Concat} do not satisfy associative property.

Besides, the global skip connection, which is the structure of adding input images to the output of the network, has been applied a lot to the latest deep learning-based image restoration methods. The deep network can estimate residual image by using a global skip connection which contributes to improved performance. Inspired by this, we proposed two terminal codes \textbf{Terminal1} and \textbf{Terminal2}. In current layer index $l$, \textbf{Terminal1} takes the output feature of layer \text{$l-1$} as input, and conducts \textbf{Convolution} operation with output channel 3, which is image channel. If the spatial dimension of the output of the layer \text{$l-1$} is not the same as that of the desired output, the spatial dimension of the feature is matched to be the same with desired size by using DMM. \textbf{Terminal2} also takes the previous layer as input, and add input image to the output of \textbf{Terminal1} operation. Figure \ref{fig2} illustrated an example that our proposed NSC generates more various architecture than existing NAS approaches.

\subsection{Dimension matching module}
Because we propose various operations that modifies feature shape, such as \textbf{Downsampling}, \textbf{Upsampling} and \textbf{Concat}, the features of each layer may have different shapes. \textbf{Elemental Add}, \textbf{Concat}, \textbf{Terminal1} and \textbf{Terminal2} suffer from dimension mismatch. \textbf{Elemental Add} requires that two input tensors should have same shape and \textbf{Concat} can be operated in case that two input tensors should have same spatial size. In addition, the output of \textbf{Terminal1} and \textbf{Terminal2} must be the image for any input feature. Dimension mismatch problem is one of the crucial factor in neural architecture design. If the shape of the feature is adjusted without considering characteristic of the feature, the lack of feature diversity can be the bottleneck for further performance improvement. Although there are several dimension matching techniques, CGP-CNN and \cite{FBNetv2}, these methods are not adequate to represent detail image component and produce limitation performance in image denoising task.

Then, we propose the dimension matching module that adaptively resizes required feature $F\in \mathbb{R}^{h\times w\times c}$ into the desired output $F'\in \mathbb{R}^{h'\times w'\times c'}$. Our DMM consists of two operation for light calculation such as $C_{s}(\cdot)$ and $P_{s}(\cdot)$. $C_{s}(\cdot)$ is trainable 1$\times$1 convolution layer, which expand channel size of input tensor by $s$ times. $P_{s}(\cdot)$ is pixel-shuffle, which rearranges elements in a tensor of shape $(C\times s^2, H\times W)$ to a tensor of shape $(C,H\times s, W\times s)$. In our search space, DMM is used in three cases for \textbf{Pred2} and inputs of \textbf{Terminal1} and \textbf{Terminal2}, i.e., (a) spatial size mismatch, (b) channel size mismatch and (c) spatial and channel size mismatch. 
(a) can appear in \textbf{Elemental Add}, \textbf{Concat}, \textbf{Terminal1} and \textbf{Terminal2}. The spatial matching module $SM(\cdot)$ resizes input tensor $F$ into $\Hat{F}$ to solve (a) problem as:
\begin{equation} \label{spatial-matching}
\Hat{F}=SM(F)\left\{\begin{array}{cl}
P_{h'/h}(C_{(h'/h)^2}(F)),& \mbox{$h'/h > 1$} \\
C_{(h'/h)^2}(P_{h'/h}(F)),& \mbox{$h'/h < 1$} \\
F.& \mbox{$h'/h=1$}\end{array}\right.
\end{equation}

In \textbf{Concat}, \textbf{Terminal1} and \textbf{Terminal2}, we do not need to consider (b) and (c) problem. However, \textbf{Elemental Add} requires that \textbf{Pred1} and \textbf{Pred2} have the same dimension. To resize channel of tensor, we designed the channel matching module $CM(\cdot)$, which simply resizes input tensor as:
\begin{equation} \label{channel-matching}
\Hat{F}=CM(F)\left\{\begin{array}{cl}
F,& \mbox{$c'/c = 1$} \\
C_{c'/c}(F).& \mbox{elsewhere}\end{array}\right.
\end{equation}

When both dimensions of spatial and channel need to be adjusted, DMM applies spatial matching module and channel matching module to predecessor:
\begin{equation} \label{DMM}
\Hat{F} = CM(SM(F)).
\end{equation}

Our DMM, which is carefully designed, allows denoiser to have a flexible structure that can combine information of various features by training with denoiser and it only requires minimal computation and parameters.

\begin{algorithm}[tb]
\caption{Search process}
\label{search}
\textbf{Input}: Train datasets, Validation datasets and Agent\\
\begin{algorithmic}
\STATE Let $t=0$.
\WHILE{not converged}
\STATE NSC = Agent($t$)
\STATE Generate denoising block $f(\cdot)$ by NSC
\STATE Train $f(\Phi^{\mathrm{T}}(\cdot))$ using Eq.~\eqref{eq11} with early stop strategy
\STATE Evaluate $f(\Phi^{\mathrm{T}}(\cdot))$ using validation datasets
\STATE Calculate reward using Eq.~\eqref{eq9} and \eqref{eq10}
\STATE Update Q-value and replay memory of Agent
\STATE t = t + 1
\ENDWHILE
\end{algorithmic}
\end{algorithm}

\subsection{Designing deep network denoiser with reinforcement learning}

We employed a denoising prior-based image restoration algorithm as a reliable outer architecture in the image restoration, which is designed by stacking \text{$K$} identical denoising blocks $f(\cdot)$. It can be represented by combining \eqref{eq4} and \eqref{eq5} as:
\begin{equation} \label{total_eq}
x^{(k+1)} = \Hat{\Phi} x^{(k)} + \delta \Phi^\mathrm{T} y + \delta\eta f(x^{(k+1)}).
\end{equation}

As proven in references, "well-formed" denoising block can improve the performance of the overall model. Therefore, we only need to search and evaluate one denoising block $f(\cdot)$ to find the structure instead of training and evaluating the entire model. The block-wise design with a reliable outer structure achieves high performance and also has good generalization ability to various datasets and applications. As our proposed DPNAS searches and evaluates one block, it leads to extremely reduced time and required the number of GPUs.

Although employing a denoising prior image restoration algorithm allows the search time to be extremely compressed, we still have to find the optimal one out of a huge number of network structures. To find the denoising block architecture efficiently, we employ Q-learning that is a popular reinforcement algorithm that aims at selecting an action that maximizes the cumulative reward.

\begin{figure*}[htbp] 
  \centering
  \includegraphics[width=1\textwidth]{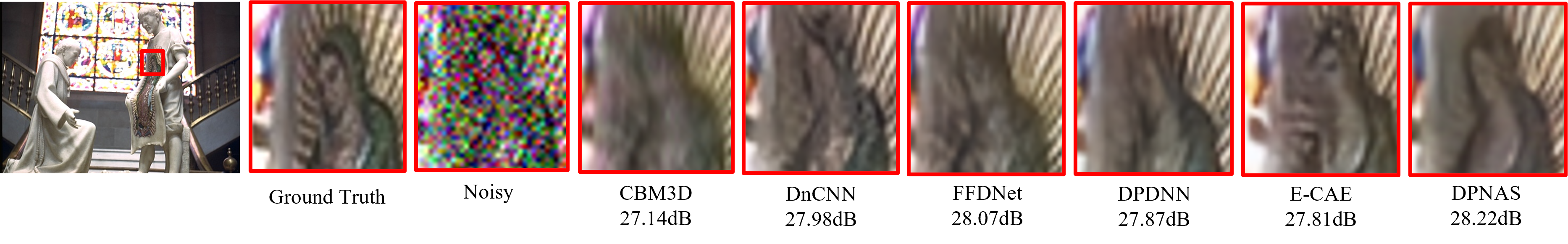}
  \caption{A synthetic noise removal example for comparison DPNAS against competitive algorithms. The value is PSNR.}
  \label{synthe}
\end{figure*}

\begin{table*}[htbp]
\centering
\begin{tabular}{ c | c || c  | c |c |c | c  || c |c| c }
\hline
Dataset &
Noise Level &
CBM3D &
DnCNN &
FFDNet &
MemNet &
DPDNN &
E-CAE &
HiNAS &
DPNAS \\
\hline\hline
          & \text{$\sigma = 30$} & 29.73 &  \textit{\underline{30.40}} & 30.31 & 28.39 & 30.37 & 30.25 & 30.09 & \textbf{30.49}\\ 
CBSD 68   & \text{$\sigma = 50$} & 27.37 &  \textit{\underline{27.97}} & 27.96 & 26.33 & 27.96 & 27.80 & 27.78 &\textbf{28.14}\\ 
          & \text{$\sigma = 70$} & 26.00 &  26.56 & 26.53 & 25.08 & \textit{\underline{26.70}} & 26.33 & 26.45 & \textbf{26.81}\\ 
\hline
          & \text{$\sigma = 30$} & 30.89 &  31.39 & 31.39 & 29.67 & \textbf{31.59} & 31.37 & 31.12 & \textit{\underline{31.55}} \\ 
Kodak 24  &\text{$\sigma = 50$}  & 28.63  & \textit{\underline{29.16}} & 29.1  & 27.65 & \textbf{29.25} & 28.95 & 28.94 & \textbf{29.25}\\ 
          & \text{$\sigma = 70$} & 27.27  & 27.64 & 27.68 & 26.40 & \textbf{28.07} & 27.47 & 27.36 &\textit{\underline{27.89}}\\ 

\hline
\end{tabular}
\caption{Quantitative results (PSNR) about color image denoising.}
\label{table3}
\end{table*}



The performance of reinforcement learning is highly dependent on how rewards are designed. If only the PSNR value that is evaluation result of validation sets is simply set as a reward, NAS model may generate an overfitting architecture for validation sets. We define the reward function as 

\begin{equation} \label{eq9}
reward = r_L = \textmd{PSNR}_{EarlyStop} - \mu \textmd{log}(\textmd{Param.}),
\end{equation}
where Param. denotes the number of trainable parameters for the searched architecture. \text{$\mu$} is a hyperparameter that controls how lightly to construct the model. It is important to appropriately set the value of \text{$\mu$} in order to have block with neither too few nor too many parameters. Besides, proper \text{$\mu$} value allows block architecture to be generalized model. We employed the Early Stop strategy to efficient search. We stop the train when the performance of searched model don't increase in a predefined interval. The PSNR\text{$_{EarlyStop}$} is the result that is the highest PSNR value during all evaluation. Unlike common NAS methods, all of the layer without successor in the searched block are not activated in training and evaluation. In this paper, the intermediate reward \text{$r_l$} is defined according to the activation of the layer \text{$l$} as :

\begin{equation} \label{eq10}
r_l=\left\{\begin{array}{cl}
r_L,& \mbox{when layer \text{$l$} is activated} \\
0.& \mbox{elsewhere}\end{array}\right.
\end{equation}

The strategy that layers without successor are eliminated by block encourages agent to make efficient network architecture.

The details of our learning procedure are illustrated in Algorithm~\ref{search}. First, the agent generates a set of NSCs with epsilon-greedy strategy and architecture for one denoising block corresponding to NSCs. Secondly, we connects \text{$\Phi^\mathrm{T}$} to denoising block \text{$f(\cdot)$}, and train the module \text{$f(\Phi^\mathrm{T}(\cdot))$} by using the loss function as

\begin{equation} \label{eq11}
Loss_{search} = ||x-f(\Phi^{\mathrm{T}}(y))||^2_2.
\end{equation}

Lastly, the reward is calculated using the PSNR value for validation sets and the number of parameter of denoising block according to Eq.~\ref{eq9} and Eq.~\ref{eq10}, and these are stored in replay memory. The agent extracts 64 block structures and their rewards from the memory, the Q-value is updated within a predefined interval. After enough training the Q-value of the agent, we select one of the architectures that are picked by the agent and insert the denoising block to denoising prior-based structure in $f(\cdot)$. The entire model consists of \text{$K$} modules with denoisers, which do not share the weights, and is trained by minimizing loss function as

\begin{equation} \label{eq12}
Loss_{DP} = ||x-x^{(K)}||^2_2,
\end{equation}
where \text{$x^{(K)}$} is final output of Eq.~\ref{total_eq}.

\section{Experiments}
In this section, we implemented two experiments, such as synthetic noise removal and real noise removal, to prove the effectiveness of the proposed method. Moreover, we present ablation study to verify utility of our search space and DMM. In the last section, we analyze the denoising block generated by proposed method. 

In image denoising task, degradation operation \text{$\Phi$} and \text{$\Phi^\mathrm{T}$} are considered as \text{$\Phi=\Phi^\mathrm{T}=I$}, where \text{$I$} is identity matrix. The hyperparameter \text{$\mu$} that controls the reward function is 0.5. We set the max layer index to 15. The number of the sampled blocks is approximately 3,000. For learning the parameters of an optimal architecture, we set \text{$\delta$} and \text{$\eta$} in denoising prior-based model to 0.1 and 0.9, respectively. The entire network training is performed for 300 epochs and ADAM optimization is used. The learning rate is initialized at \text{$10^{-3}$} and decreases by half every 50 epochs. To train the model, we randomly extract 64\text{$\times$}64 image patches from training images, and use a batch size of 64. We use images with center crop as validation images from validation set. The searching and training process takes approximately 1 day using a RTX 3090 GPU. The experiments for other datasets and details not included in this paper can be found in the supplementary material.

\subsection{Synthetic noise removal}
DPNAS is searched and evaluated for AWGN noises of different levels (e.g., 30, 50, and 70). Training dataset is 800 DIV2K training images, and validation dataset is 100 DIV2K validation images. We compare our DPNAS with state-of-the-art color denoising methods CBM3D~\cite{BM3D}, DnCNN, FFDNet~\cite{FFDNET}, MemNet~\cite{memnet}, DPDNN~\cite{denoisingprior}, E-CAE~\cite{ECAE} and HiNAS. DPDNN is the baseline of our DPNAS, and E-CAE and HiNAS is denoising model generated by state-of-the-art NAS. For fair comparisons, we re-implement the methods using the code provided by the authors under the identical implementation settings indicated at the beginning of this main section. Because the public code of HiNAS is not provided, we reproduce HiNAS based on paper with our best.

Table~\ref{table3} presents PSNR results for the denoising of BSD 68~\cite{BSD68} and Kodak 24 datasets with the best performance marked in bold and the second best performance is italicized. The models that are generated from DPNAS produce superior performance compared with state-of-the-art methods and outperform E-CAE and HiNAS with a large margin for all noise levels. The E-CAE spends four days with four GPUs to search the network architecture, but our DPNAS found the denoising block architecture in one day with one GPU. Figure~\ref{params} lists the number of network parameters including models generated by DPNAS under noise levels 30, 50, and 70, respectively. The proposed DPNAS models usually require fewer parameters than existing models. Figure~\ref{synthe} presents qualitative results of DPNAS in synthetic noise removal with noise level 50. The zoomed results demonstrate a superior visual quality and restoration performance.

\begin{figure*}[htbp] 
  \centering
  \includegraphics[width=1\textwidth]{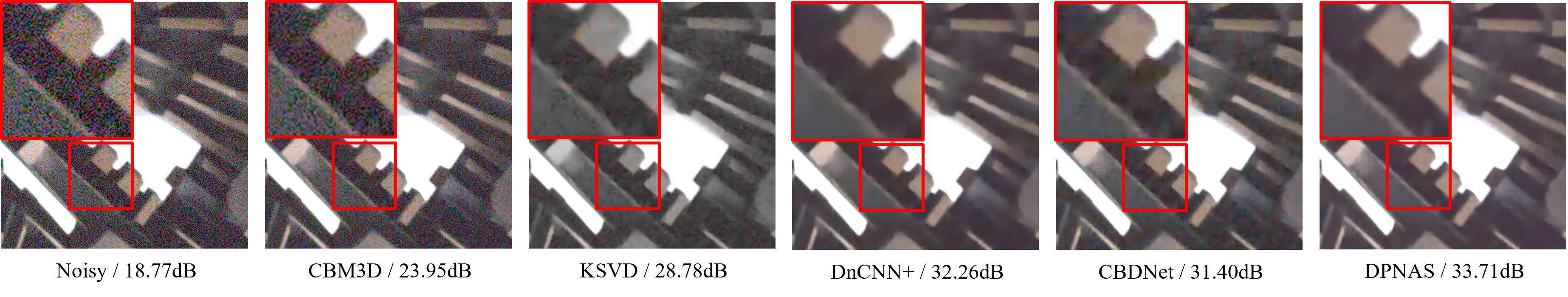}
  \caption{A real noisy example from DnD dataset for comparison DPNAS against competitive algorithms. The value is PSNR.}
  \label{fig5}
\end{figure*}

\begin{table}[t]
\centering
\begin{tabular}{ l || c  c  c }
\hline 
 
 Methods &
 Blind/Non-blind &
 PSNR &
 SSIM \\
\hline\hline
  FFDNet    &  Non-Blind & 34.40 &  0.8474  \\
  CBM3D     &  Non-Blind & 34.51 &  0.8507  \\
  KSVD      &  Non-Blind & 36.49 &  0.8978  \\
  FFDNet+   &  Non-Blind & 37.61 &  0.9415  \\
  DnCNN+    &  Non-Blind & 37.90 &  0.9430  \\
  CBDNet    &  Blind     & 38.06 &  0.9421  \\
  DPNAS &  Blind & \textbf{38.96} &  \textbf{0.9476}  \\
  
\hline
\end{tabular}
\caption{The Mean PSNR and SSIM denoising results on the DnD sRGB images}
\label{table5}
\end{table}

\subsection{Real noise removal}
We employ a real noise dataset, in which the noise is spatially variant and correlates with the image, to demonstrate the practicality of the proposed DPNAS. We train and test generated architecture using the reliable DnD real noise dataset~\cite{CBDNet}. To find the denoising block, we randomly choose 5\text{$\%$} of DnD training images as the validation set. The rest of the dataset is used as the training dataset for searched blocks. In Table~\ref{table5}, 'non-blind' represents the results of training and testing the models using dataset with the noise level, while 'blind' denotes the result without the noise level. Table~\ref{table5} shows the quantitative results for sRGB data in the DnD dataset using existing methods and models generated by DPNAS. Our DPNAS presents better performance than the other compared method with large margin. From the perspective of visual quality, we show the results of various methods in Figure~\ref{fig5}. Most of the compared denoising methods either cannot remove noise or produce some artifacts. On the other hand, the magnified red boxes restored by other denoising algorithms are over-smoothed. In contrast, our DPNAS obtains results that maintained the shape of the structure of image.

\subsection{Ablation study}
In the ablation study, we perform two comparison experiments. First, we analyze the performance and search time depending on whether our search space and DMM are used. Secondly, we demonstrated the generalization performance of the searched block with the constrained reward with the number of parameters. In this experiment, we set AWGN removal with a noise level of \text{$\sigma$}=50. Training details are the same as one of section \textbf{Synthetic noise removal}.

\subsubsection{The comparison of the performance and search time according to search space and DMM}
Table~\ref{ablation} presents the performance and search time in each case. Case1 is DPNAS with search space of Block-QNN instead of our proposed search space. Case2 and Case3 are DPNAS without proposed DMM and DPNAS with dimension matching strategy of CGP-CNN, respectively. In Case1, finally generated architecture doesn't select max-pooling and avg-pooling, and only contains $3\times3$ convolution layers. It implies that the search space for image denoising is needed in image denoising task. Since Case2 causes dimension mismatch of network, network search processing has long search time. Case3 has low performance and longer search time than DPNAS, because conventional dimension matching strategy do not adequate in image denoising task, which reconstructs detail components. The proposed DMM, which is carefully designed, can effectively solve the dimension mismatch problem from NAS.

\begin{table}[t]
\begin{center} 
{\footnotesize
\begin{tabular}{|c||c|c|c|c|c|}
\hline
Case  & Case1 & Case2 & Case3 & DPNAS  \\
\hline\hline
PSNR                & 26.84  & 27.99  & 28.16 & 28.51  \\
\hline
Search time (h)     & 8      & 34     &22     & 18     \\

\hline
\end{tabular}
}
\end{center}
\caption{Comparison of the performance and search time according to modification of search space and dimension matching module.}
\label{ablation}
\end{table}

\begin{table}[t]
\begin{center} 
{\footnotesize
\begin{tabular}{|c||c|c|c|}
\hline
Datasets  & DIV2K 100 & CBSD68 & Kodak24  \\
\hline\hline
DnCNN     & 28.25     & 27.97 & 29.16    \\
DPNAS w/o & 28.65     & 27.99 & 28.96    \\
DPNAS     & 28.51     & 28.14 & 29.25    \\
\hline
\end{tabular}
}
\end{center}
\caption{Comparison of the performance between searched models with redefined reward (DPNAS) and only PSNR reward (DPNAS w/o)}
\label{table2}
\end{table}

\subsubsection{Effectiveness of constrained parameter architecture as reward}
We notice that the search process without constrained reward for the number of parameters can lead to generating overfitted network architecture, which only achieves great performance in similar images to validation domain. Table~\ref{table2} summarizes the denoising results for each datasets. DPNAS is the optimal block architecture that is searched with our redefined reward, which reduces the number of parameters. DPNAS w/o is searched block by using reward with only PSNR. Since Q-value of the agent is updated using DIV2K 100 as the validation set for evaluation of searched architecture, DPNAS and DPNAS w/o outperform the DnCNN. However, DPNAS w/o shows poor generalization performance on other test sets. In contrast, DPNAS outperforms compared denoising method on test sets as well as DIV2K 100. This experiment demonstrates that our redefined reward encourages the agent to find generalized denoising block architecture.

\begin{figure}[t] 
  \centering
  \includegraphics[width=0.5\textwidth]{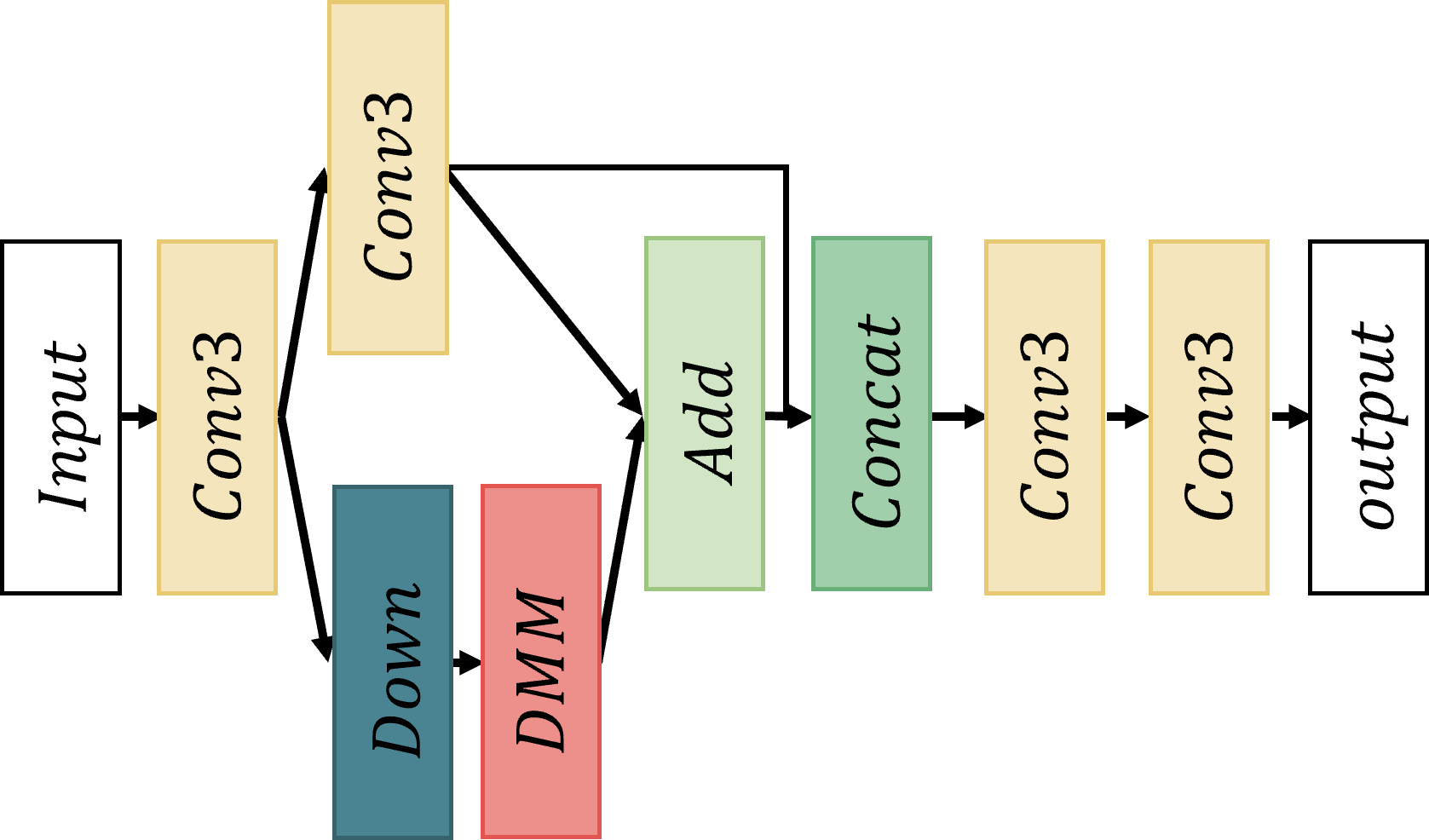}
  \caption{The denoising block architecture on noise level 30.}
  \label{fig6}
\end{figure}

\subsection{Denoising block architecture analysis}

In this section, we analyze the architecture generated by DPNAS for synthetic noise removal model with noise levels 30 as shown in Figure \ref{fig6}. We can see that denoising block has the two-path architecture. The under path is bottleneck structure, and the other is residual and skip-connection structure. The denoising blocks tend to have the deeper bottleneck structure when noise intensity is stronger, because bottleneck structure generally is effective in eliminating redundant elements. The detail comparison of generated networks is illustrated in our supplementary materials.

\subsection{The figures according to main manuscript}

\begin{figure*}[t] 
  \centering
  \includegraphics[width=1\textwidth]{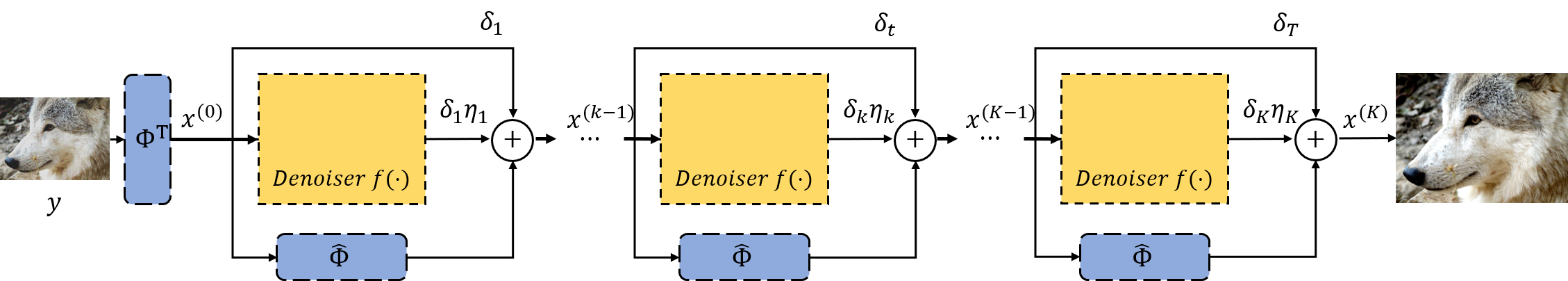}
  \caption{Framework of denoising prior-based architecture with a fixed number of modules $K$.}
  \label{fig_overall}
\end{figure*}
\begin{figure*}[t] 
  \centering
  \includegraphics[width=1\textwidth]{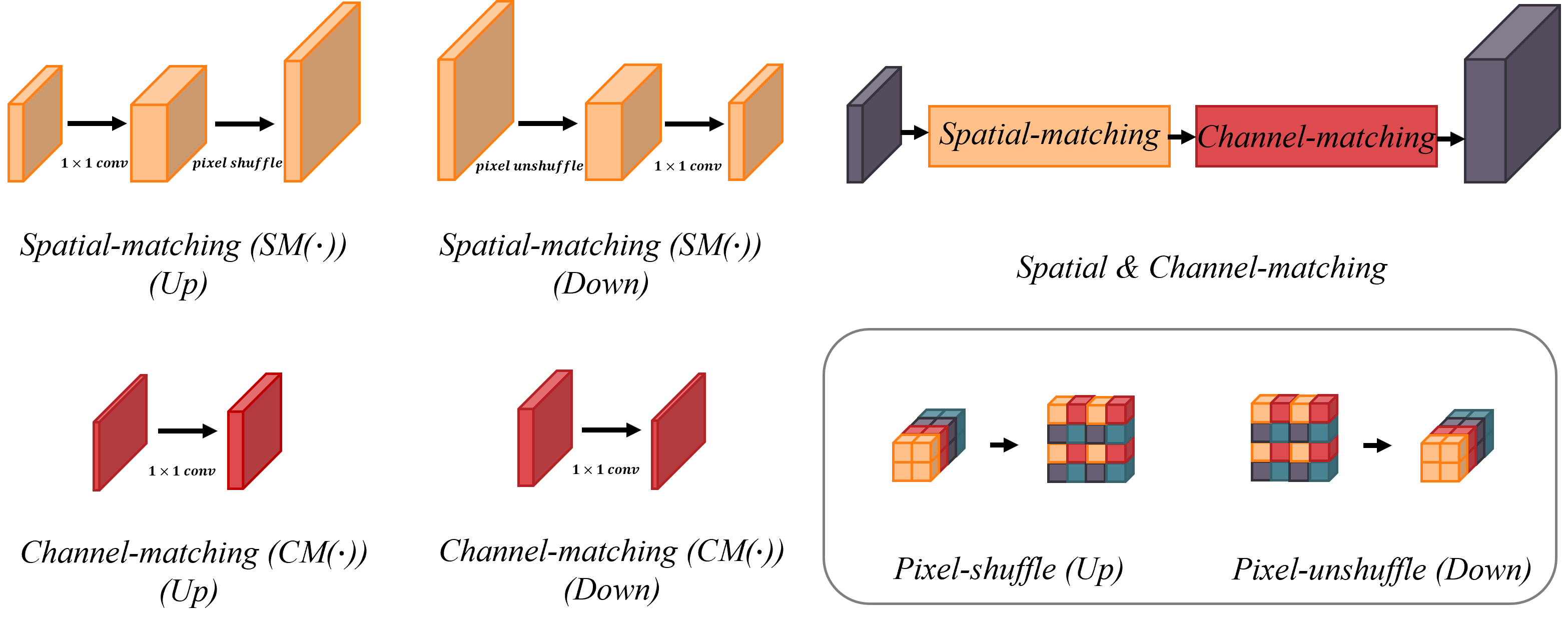}
  \caption{The visual expression of the proposed dimension matching module.}
  \label{fig_DMM}
\end{figure*}

Figure~\ref{fig_overall} shows the overall architecture of Eq.(9) in main manuscript. Figure~\ref{fig_DMM} shows the proposed dimension matching module architectures of Eq.(6), (7) and (8) in main manuscript.

\subsection{Epsilon greedy strategy}
The epsilon greedy strategy selects the best layer for a proportion \text{$1-\epsilon$} of the trials, and the layer is selected at random with uniform probability for a proportion \text{$\epsilon$}. We train the agent with 100 iterations while sampling 3000 blocks. The \text{$\epsilon$} is initialized at 1 and decreases smoothly to 0.1 as shown in Table~\ref{table7}. It allows the agent to transform from exploration to exploitation.

\begin{table}[htbp]
\begin{center} 
{\footnotesize
\begin{tabular}{|c|c|c|c|c|c|}
\hline
\text{$\epsilon$}  & 1.0 &  0.9 & 0.8 & 0.7 & 0.6 \\
\hline
Iters & 50& 5 & 5& 5& 5       \\
\hline
\hline
-  & 0.5 & 0.4 & 0.3 & 0.2 & 0.1\\
\hline
- & 5& 5& 5& 5& 10  \\
\hline

\end{tabular}
}
\end{center}
\caption{The epsilon schedule. Iters denotes the number of iteration that the agent is trained with \text{$\epsilon$}.}
\label{table7}
\end{table}

\begin{figure}[t] 
  \centering
  \includegraphics[width=0.50\textwidth]{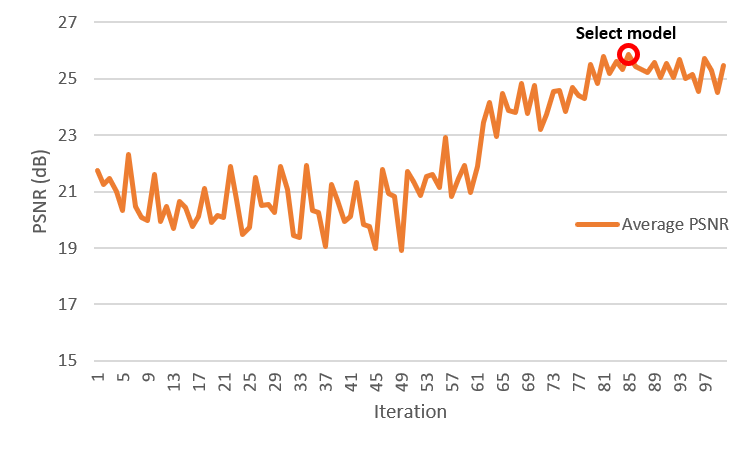}
  \caption{Q-learning performance. PSNR result according to the number of training iteration.}
  \label{fig_select}
\end{figure}

\subsection{Q-learning performance and selected network architecture}
In this section, we show performances for the validation set according to iteration as shown in Figure~\ref{fig_select}. Training details are the same as Section 4.1 Ablation study in the main paper. When the epsilon \text{$\epsilon$} decreases, the agent can take greedy action and generates better denoising architecture than random searching. We chose network architecture for denoiser with the best PSNR during the last 20 epochs in all experiments.

\subsection{PSNR comparison with various number of \text{$K$}}
\begin{figure}[t]
  \centering
  \includegraphics[width=0.5\textwidth]{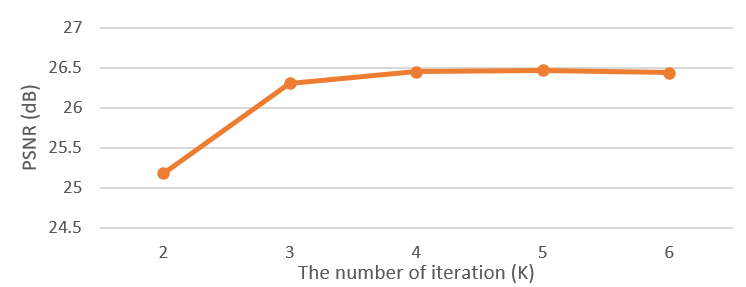}
  \caption{PSNR comparison according to the number of denoising block \text{$K$}} \label{fig4}
\end{figure}

Figure~\ref{fig4} illustrates the average PSNR curves for the denoising results on the 100 validation images (DIV2K validation sets) based on the number of denoising block \text{$K$}. The model seems to converge at \text{$K=4$} gradually. Taking into account the balance between model complexity and denoising performance, we set the number of denoising block iteration \text{$K$} to 4 in the remaining experiments.

\subsection{The visual quality comparison of noise removal}
\begin{figure*}[t] 
  \centering
  \includegraphics[width=1\textwidth]{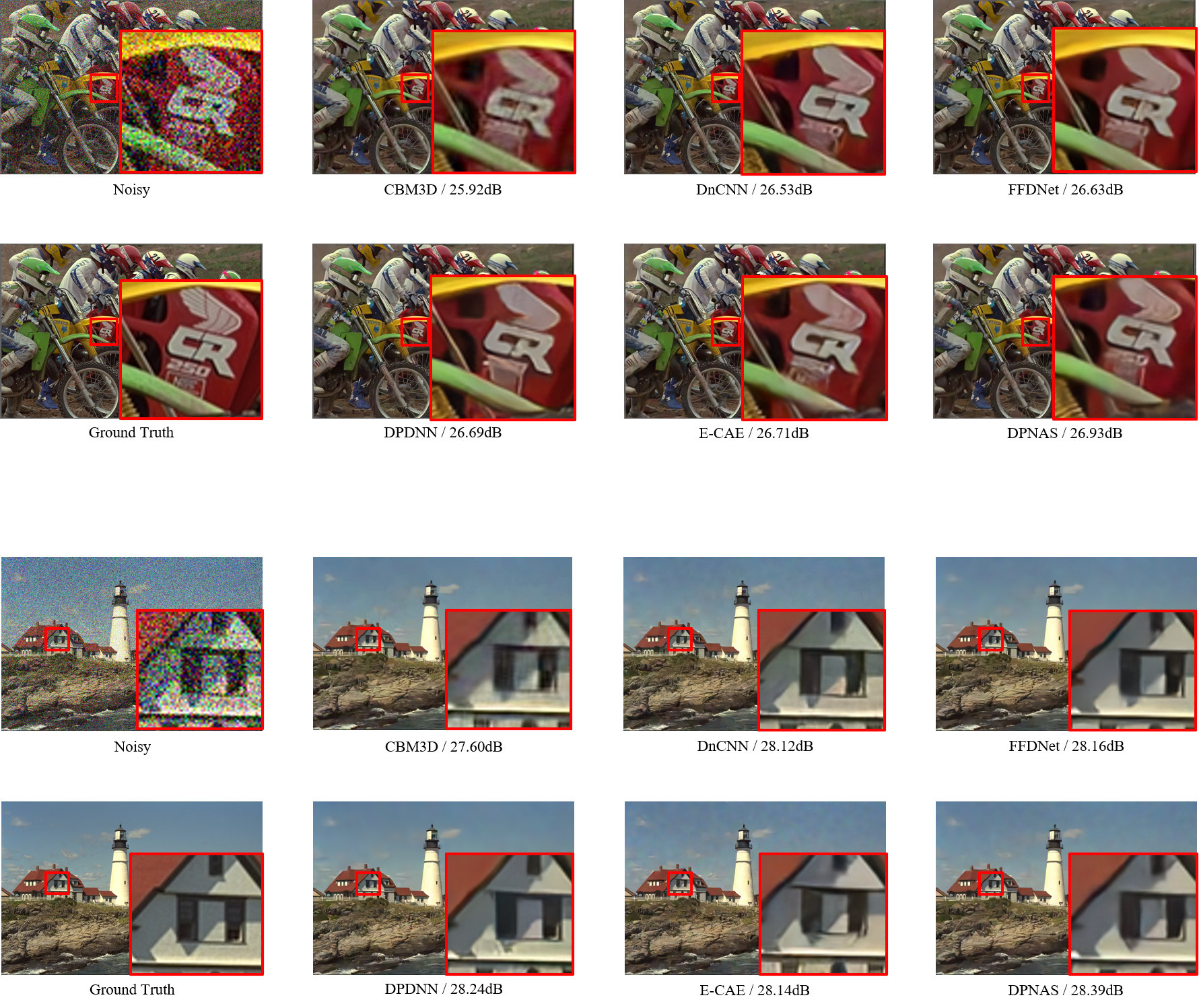}
  \caption{Additional results of qualitative comparison.}
  \label{syntheticresult}
\end{figure*}

\begin{figure*}[t] 
  \centering
  \includegraphics[width=1\textwidth]{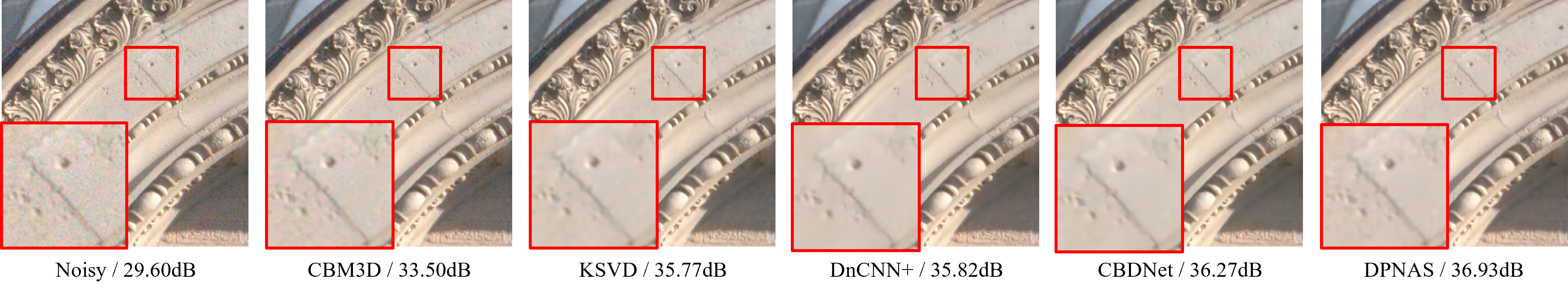}
  \caption{Additional results of qualitative comparison.}
  \label{fig_real_result}
\end{figure*}

Finally, we present extra qualitative results of DPNAS in synthetic noise removal with noise level 50 and real noise on test dataset. Figure~\ref{syntheticresult}, \ref{fig_real_result} illustrate synthetic and real noise removal results, respectively. The zoomed results demonstrate a superior visual quality and restoration performance.

\begin{figure*}[t] 
  \centering
  \includegraphics[width=1\textwidth]{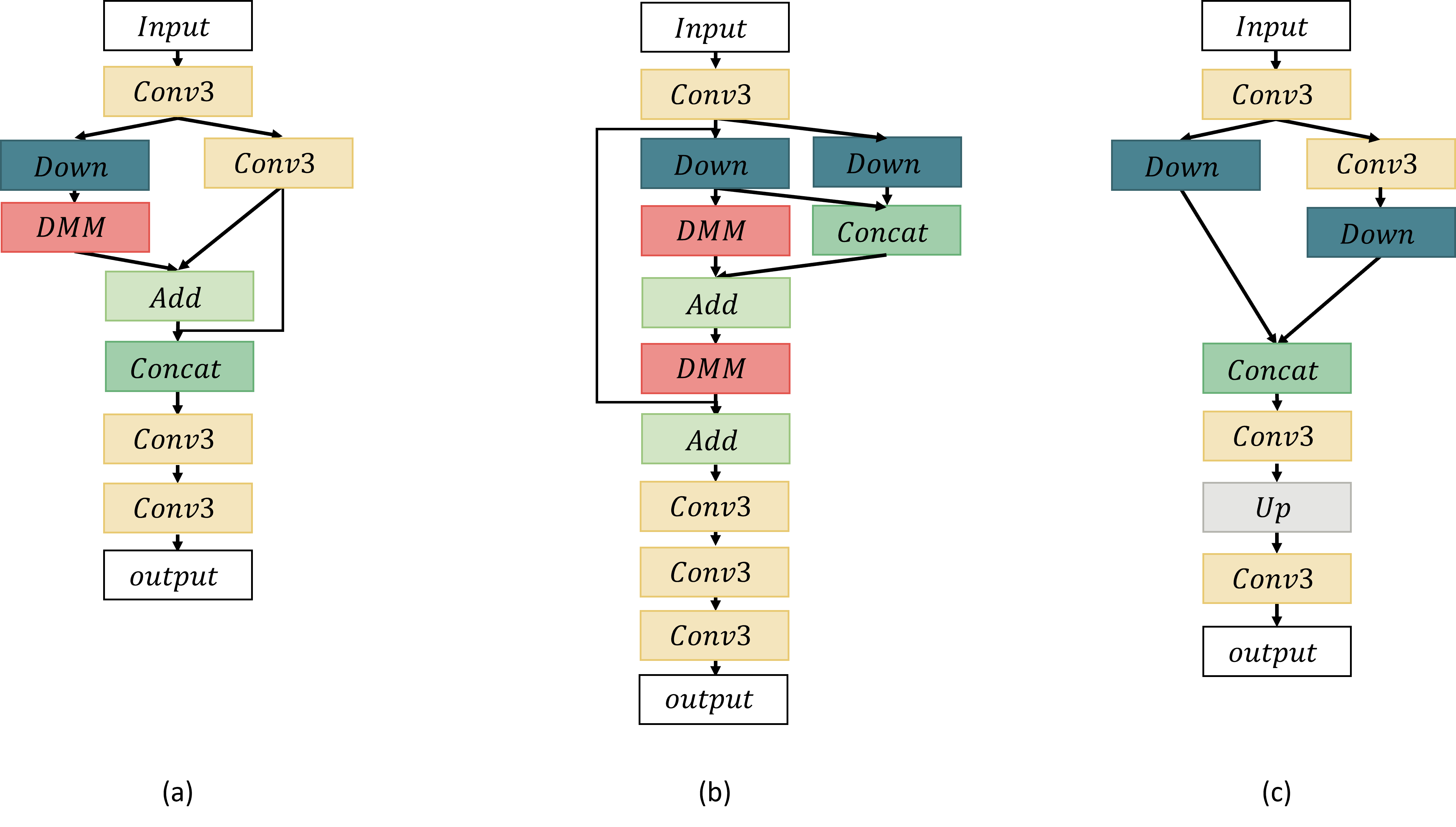}
  \caption{Additional results of qualitative comparison.}
  \label{denoising_networks}
\end{figure*}

\subsection{Denoising block architectures analysis}
In this section, we analyze the architectures generated by DPNAS and show three architecture denoising blocks for synthetic noise removal model with noise levels 30, 50, and 70, respectively as shown in Figure~\ref{denoising_networks}. We can see that the three common network structures appear in three denoising blocks. First, it is observed that the first layers of (a), (b) and (c) are defined as \textbf{Convolution} with kernel size 3, which denotes that the non-linear feature space affects the denoising performance in practice. Second, the generated models contain bottle-neck architecture by employing \textbf{Downsampling} layer, because bottle-neck structure generally is effective in eliminating redundant elements. Lastly, three denoising blocks select \textbf{Terminal1} as the last layer that consists of \text{$3\times 3$} convolution layer, our deep networks that has small number of parameters better estimate image than residual image. The denoising blocks tend to have the deeper the bottleneck structure when noise intensity is stronger.

\subsection{Topology of the real noise removal block structure}
\begin{figure}[t] 
  \centering
  \includegraphics[width=0.2\textwidth]{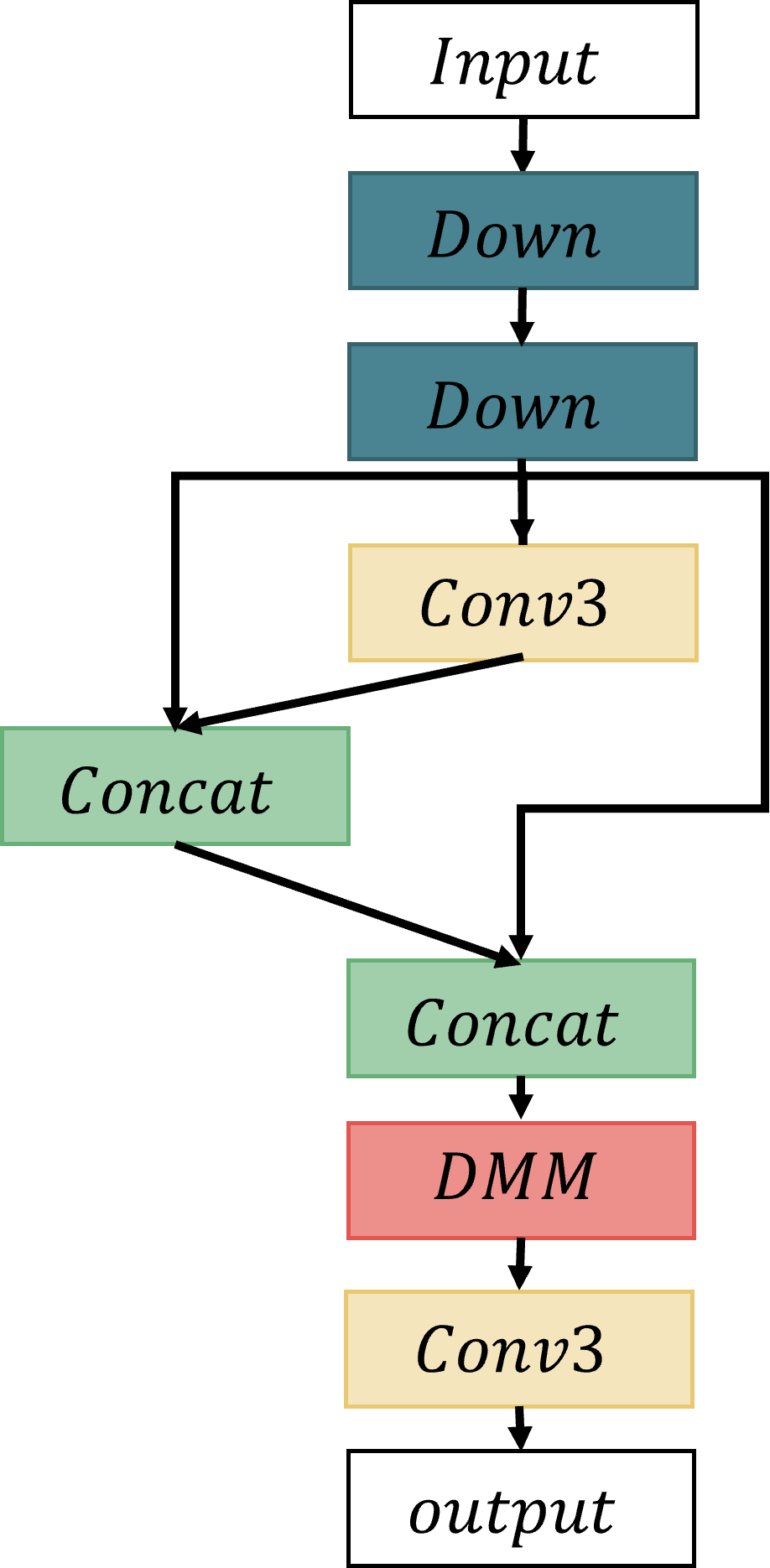}
  \caption{The denoising architecture for real noise removal.}
  \label{real_network}
\end{figure}

We already acquired impressive analysis results for several denoising block architectures on synthetic noise removal and super-resolution. Therefore, we need to analyze denoising block architecture for real noise removal, because the characteristic of real noise is different from the characteristic of synthetic noise. Figure~\ref{real_network} illustrates denoising block architecture generated by DPNAS. Similar to other denoising networks, the denoising block for real noise has a strong bottleneck structure. To compensate for detailed information loss caused by bottleneck structure, DPNAS employs channel expansion operation. Our DPNAS generated memory-efficient model architecture with 437K trainable parameters, which has around 12 times fewer parameters than CBDNet, which has 5,332K parameters.

\begin{table*}[htbp]
\centering

\begin{tabular}{ c | c  c c c  c  c c  c c c  c c }
\hline
\multirow{2}{*}{Test Set} &
\multicolumn{2}{c}{TNRD} &
\multicolumn{2}{c}{SRCNN} &
\multicolumn{2}{c}{VDSR} &
\multicolumn{2}{c}{DnCNN} &
\multicolumn{2}{c}{FALSR-B} &
\multicolumn{2}{c}{DPNAS} \\

       & PSNR  & SSIM  & PSNR  & SSIM  & PSNR  & SSIM  & PSNR  & SSIM  & PSNR  & SSIM & PSNR & SSIM\\
\hline\hline
Set 5  & 36.86 & 0.956 & 36.66 & 0.954 & 37.53 & \textbf{0.959} & 37.58 & \textbf{0.959} & \textit{\underline{37.61}} & \textit{\underline{0.958}}& \textbf{37.63} & \textbf{0.959} \\ 
Set 14 & 32.54 & 0.907 & 32.42 & 0.906 & 33.03 & \textit{\underline{0.912}} & 33.03 & 0.911 & \textbf{33.29} & \textbf{0.914}& \textit{\underline{33.10}} & \textit{\underline{0.912}}\\ 

\hline
\end{tabular}
\caption{Quantitative results about single image super-resolution.}
\label{table_sr}
\end{table*}

\subsection{Single image super-resolution}
To train the search model for single image super-resolution, we extract 40\text{$\times$}40 image patches as low-resolution inputs from DIV2K train dataset and evaluate searched model using the DIV2K validation dataset. The degradation process $\Phi$ and $\Phi^\mathrm{T}$ are set to bicubic decimation and interpolation, respectively. Since CUDA memory issue, we used convolution with 32 output channels. We compare our DPNAS with popular super-resolution methods TNRD, SRCNN~\cite{srcnn}, VDSR~\cite{VDSR}, DnCNN, and FALSR-B~\cite{FALSR}. For fair comparisons, the results of the others are directly borrowed by corresponding papers in Table~\ref{table_sr}. DPNAS achieves competitive results than representative super-resolution algorithms and models generated by other NAS~\cite{FALSR} in test data sets, Set5 and Set14.

\begin{figure}[t] 
  \centering
  \includegraphics[width=0.3\textwidth]{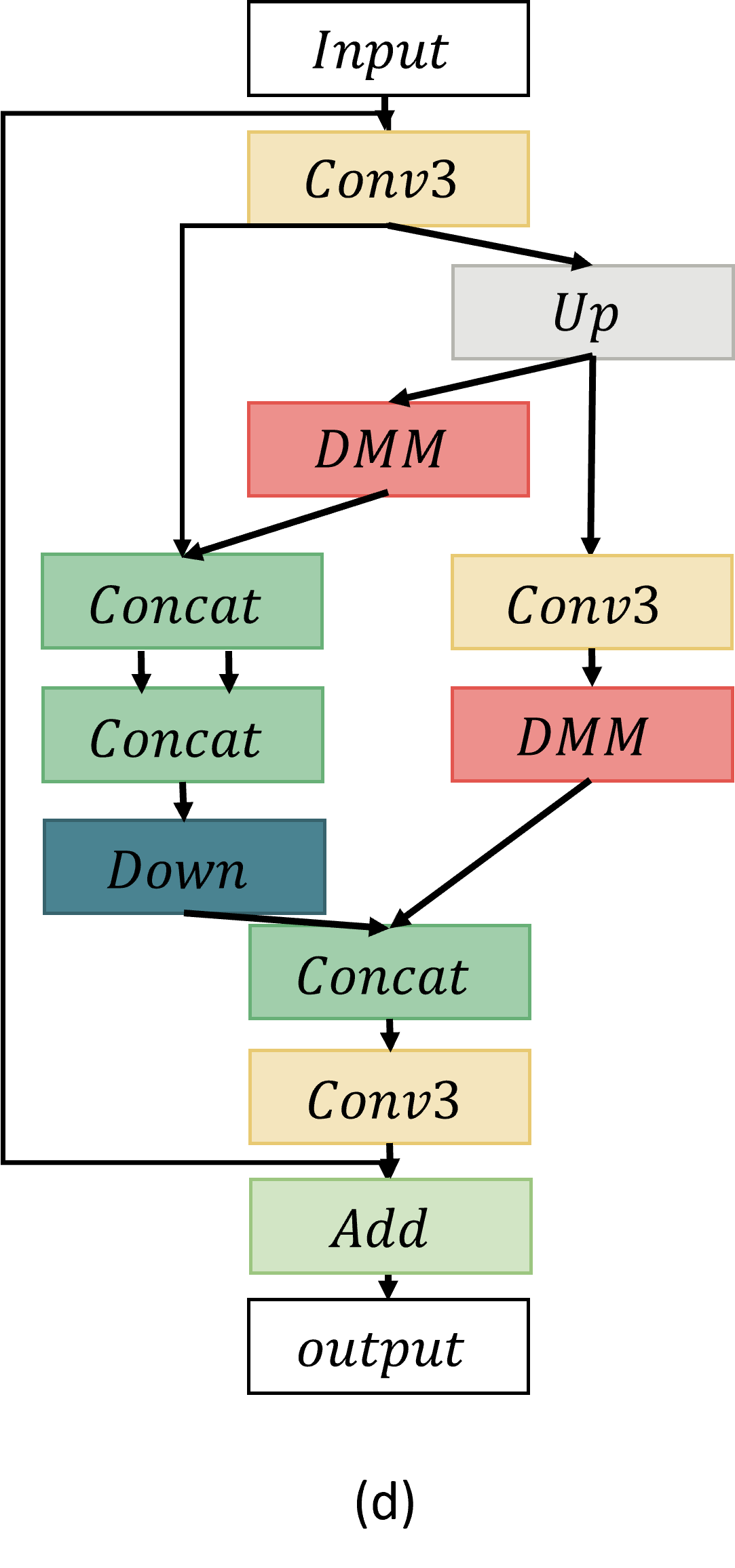}
  \caption{The denoising block architecture for single image super-resolution.}
  \label{fig_srnetwork}
\end{figure}

In block structure for super-resolution as shown in Figure~\ref{fig_srnetwork}, there are several channel expansion structures, which lead to better representation power of the restored image. Then, \textbf{Terminal2} operation consist of \text{$3\times 3$} convolution layer and element-wise add with input and is used in last layer unlike architectures for image denoising. It implies that the deep networks can better estimate the residual image corresponding observation \text{$y$} in super-resolution task.

\begin{table}[htbp]
\begin{center} 
{\footnotesize
\begin{tabular}{|c|c|c|}
\hline
Methods & PSNR & SSIM \\
\hline
LP & 20.46 & 0.7297       \\
DetailsNet & 21.16 & 0.7320 \\
JORDER & 22.24 & 0.7763  \\
JORDER-R & 22.29 & 0.7922  \\
RESCAN & 24.09 & 0.8410  \\
HiNAS & 26.31 & 0.8685 \\
DPNAS & \textbf{26.55} & \textbf{0.8702} \\
\hline

\end{tabular}
}
\end{center}
\caption{The deraining results.}
\label{derain}
\end{table}

\subsection{Single image deraining}
We applied DPNAS on a challenging deraining dataset (Rain800). We compare our DPNAS with popular deraining algorithms LP~\cite{LP}, DetailsNet~\cite{DetailNet}, JORDER~\cite{Jorder}, JORDER-R, RESCAN~\cite{rescan} and HiNAS~\cite{HINAS}. Table~\ref{derain} shows the results of deraining.

\section{Conclusion}

In this paper, we proposed a novel architecture search method for designing image denoising algorithms by identifying the component cell structure efficacious in rendering the overall network effective in its task. A set of algorithms were integrated for ensuring that the tensor dimensions are matched when designing CNN operations within the cell structure. These algorithms allowed to freely integrate a variety of combinations of CNN operations within the cell block for finding optimal designs. By implementing cell based search and the dimensionality matching algorithms, the search becomes highly efficient that it completed an architecture search for an image denoising task by just one day with a single GPU. The architecture designed by the proposed DPNAS outperformed state-of-the-art methods in synthetic noise removal and real noise removal.

\bibliographystyle{IEEEbib.bst}
\bibliography{main.bib}

\end{document}